\documentclass[11pt]{article}

\usepackage[final]{acl/acl}

\usepackage{tikz}
\usepackage{times}
\usepackage{latexsym}
\usepackage{multirow}
\usepackage{booktabs}
\usepackage{amsmath} 
\usepackage{amssymb}
\usepackage{makecell}
\usepackage{grffile}
\usepackage{enumitem}
\usepackage{tcolorbox}
\usepackage{setspace}
\usepackage{listings}
\tcbuselibrary{breakable}

\usepackage{tabularx}
\usepackage{booktabs}
\usepackage[T1]{fontenc}

\usepackage[utf8]{inputenc}

\usepackage{microtype}

\usepackage{inconsolata}

\usepackage{graphicx}
\usepackage{footnote}
\definecolor{tokengreen}{HTML}{66CC66}

\newcommand{\perf}[2]{#1$_{/\text{\footnotesize\color{tokengreen}#2}}$}

\title{Breaking Contextual Inertia: Reinforcement Learning with Single-Turn Anchors for Stable Multi-Turn Interaction}

\author{
  Xingwu Chen\textsuperscript{1}, 
  Zhanqiu Zhang\textsuperscript{2,$\dagger$}, 
  Yiwen Guo\textsuperscript{3,$\dagger$}, 
  Difan Zou\textsuperscript{1,$\dagger$} \\
  School of Computing and Data Science, The University of Hong Kong \textsuperscript{1} \\
  LIGHTSPEED\textsuperscript{2} \\
  Independent Researcher\textsuperscript{3} \\
  \texttt{xingwu@connect.hku.hk}, \texttt{zqzhang27@gmail.com}, \\
  \texttt{guoyiwen89@gmail.com}, \texttt{ dzou@cs.hku.hk} 
}

\begin{document}
\maketitle
{\renewcommand{\thefootnote}{}\footnotetext{$^\dagger$ Corresponding authors.}}
\begin{abstract}

While LLMs demonstrate strong reasoning capabilities when provided with full information in a single turn, they exhibit substantial vulnerability in multi-turn interactions. Specifically, when information is revealed incrementally or requires updates, models frequently fail to integrate new constraints, leading to a collapse in performance compared to their single-turn baselines. We term the root cause as \emph{Contextual Inertia}: a phenomenon where models rigidly adhere to previous reasoning traces. Even when users explicitly provide corrections or new data in later turns, the model ignores them, preferring to maintain consistency with its previous (incorrect) reasoning path. To address this, we introduce \textbf{R}einforcement \textbf{L}earning with \textbf{S}ingle-\textbf{T}urn \textbf{A}nchors (\textbf{RLSTA}), a generalizable training approach designed to stabilize multi-turn interaction across diverse scenarios and domains. RLSTA leverages the model's superior single-turn capabilities as stable internal anchors to provide reward signals. By aligning multi-turn responses with these anchors, RLSTA empowers models to break contextual inertia and self-calibrate their reasoning based on the latest information. Experiments show that RLSTA significantly outperforms standard fine-tuning and abstention-based methods. Notably, our method exhibits strong cross-domain generalization (e.g., math to code) and proves effective even without external verifiers, highlighting its potential for general-domain applications.
Code is available at \url{https://github.com/Tencent/RLSTA}.

\end{abstract}

\section{Introduction}

\begin{figure}[t]
\centering
\includegraphics[width=\linewidth]{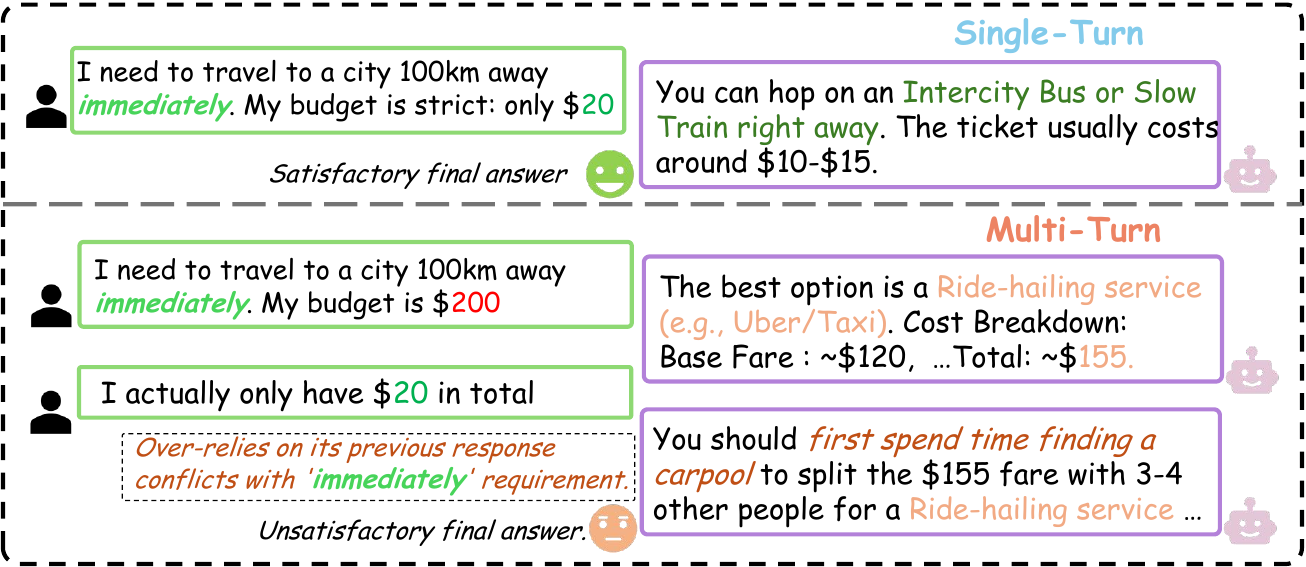}
\vspace{-20pt}
\caption{Contextual Inertia in multi-turn interaction: The persistence of the initial response leads to an unsatisfactory final answer.}
\label{fig:illustration} 
\vspace{-20pt}
\end{figure}

Large Language Models (LLMs) have demonstrated remarkable capabilities in solving complex reasoning tasks \citep{yang2025qwen3, google2025gemini3}. However, their success has predominantly been confined to single-turn settings \citep{yang2025qwen3} or scenarios involving carefully engineered inference procedures \citep{shao2025deepseekmath}. Yet, multi-turn interaction has emerged as the ubiquitous paradigm for human-AI engagement, serving as the backbone for applications ranging from general-purpose chat systems \citep{openai2025gpt5, google2025gemini3} to complex agentic workflows \citep{liu2025advances, wei2025webagent}. In these practical scenarios, users frequently introduce new conditions or correct previous requirements. However, models frequently fail to refresh their reasoning in multi-turn contexts, yielding unsatisfactory answers compared to single-turn scenarios. For instance, as illustrated in Figure~\ref{fig:illustration}, when the user updates the budget constraint, the model rigidly adheres to its previous ``ride-hailing'' plan and suggests spending time finding a carpool, producing a solution that directly violates the urgency requirement. This deficiency severely limits LLM utility in complex applications like interactive reasoning \citep{wu2025collabllm} and collaborative problem-solving \citep{chen-etal-2024-comm}.


\citet{laban2025llms} refer to this performance degradation in multi-turn settings as getting ``Lost in Conversation'' (LiC). They focus on a specific scenario where a complete condition is fragmented into multiple shards and provided to the LLM sequentially (\texttt{MT-Add}). Through large-scale experiments, they reveal that models are vulnerable in such scenarios, as LLMs often generate premature answer attempts in early turns without full information. These premature attempts can adversely affect follow-up responses, ultimately causing the model to lose track of the logical flow.

Existing approaches primarily address the symptoms of multi-turn degradation rather than investigating its root cause. One prominent line of research seeks to enhance multi-turn capabilities through direct fine-tuning strategies \citep{sun2024parrot, zhou2024archer, zhou2025sweetrltrainingmultiturnllm}. While effective for general instruction following and credit assignment, these methods rely heavily on external supervision, thereby bypassing the intrinsic degradation mechanisms rather than rectifying the model's internal failure modes. Another direction investigates behavioral strategies, such as prompting clarification requests \citep{wu2025collabllm} or encouraging active abstention \citep{li2025verifiable}, particularly when user input is insufficient. Although these methods mitigate error accumulation from premature responses, they are incompatible with scenarios necessitating dynamic state updates \citep{kwan-etal-2024-mt,bai-etal-2024-mt}, such as the \texttt{MT-Refine} setting, where the model must iteratively correct its initial response rather than remaining silent.

In this paper, we identify the \textbf{indiscriminate nature} of the model’s multi-turn behavior and \textbf{quantitatively attribute} the root cause of multi-turn interaction failures to a phenomenon we term \emph{Contextual Inertia}. Specifically, LLMs in multi-turn dialogues tend to rigidly adhere to previously generated reasoning traces, even when these traces have been explicitly negated or corrected by subsequent user information. This indiscriminate nature causes erroneous or misleading intermediate reasoning to be continuously inherited and reinforced throughout the multi-turn iteration, ultimately leading to incorrect final answers. Our analysis reveals that over 70\%--90\% of multi-turn errors can be directly traced back to the propagation of errors from previous turn responses, rather than independent reasoning failures in the final turn. Therefore, breaking such inertia and pushing the model to correct its previous failures serves as a critical and general step toward stable multi-turn interaction.


Building on this insight, we introduce \textbf{R}einforcement \textbf{L}earning with \textbf{S}ingle-\textbf{T}urn \textbf{A}nchors (\textbf{RLSTA}), a generalizable training approach designed to break contextual inertia and stabilize multi-turn interaction. RLSTA capitalizes on the model's superior reasoning capabilities in full-information single-turn settings, treating these responses as stable internal \emph{anchors} to guide multi-turn generation via reward signals. Our contributions can be summarized as follows:

\begin{itemize}[nosep, leftmargin=10pt] 
    \item We identify the \textbf{indiscriminate nature} of \emph{Contextual Inertia} and \textbf{quantitatively attribute} its impact. We define this phenomenon as the model's indiscriminate adherence to previous reasoning traces, providing a new perspective on addressing failures in dynamic interactions.

    \item We propose Reinforcement Learning with Single-Turn Anchors (RLSTA), a generalizable training approach designed to break \emph{Contextual Inertia} by leveraging the model's superior single-turn capabilities as internal reward signals. Unlike previous methods tailored for specific tasks, RLSTA is applicable to diverse interaction scenarios, including both incremental information addition (\texttt{MT-Add}) and error correction (\texttt{MT-Refine}).

    \item We conduct extensive experiments demonstrating that RLSTA not only outperforms standard tuning methods but also achieves comparable or even superior performance compared to abstention-based fine-tuning approaches. Moreover, RLSTA  exhibits cross-domain generalization and proves effective even without external verifiers, highlighting its potential for general-domain applications.
    
\end{itemize}

\begin{figure*}[t]
\centering
\includegraphics[width=\linewidth]{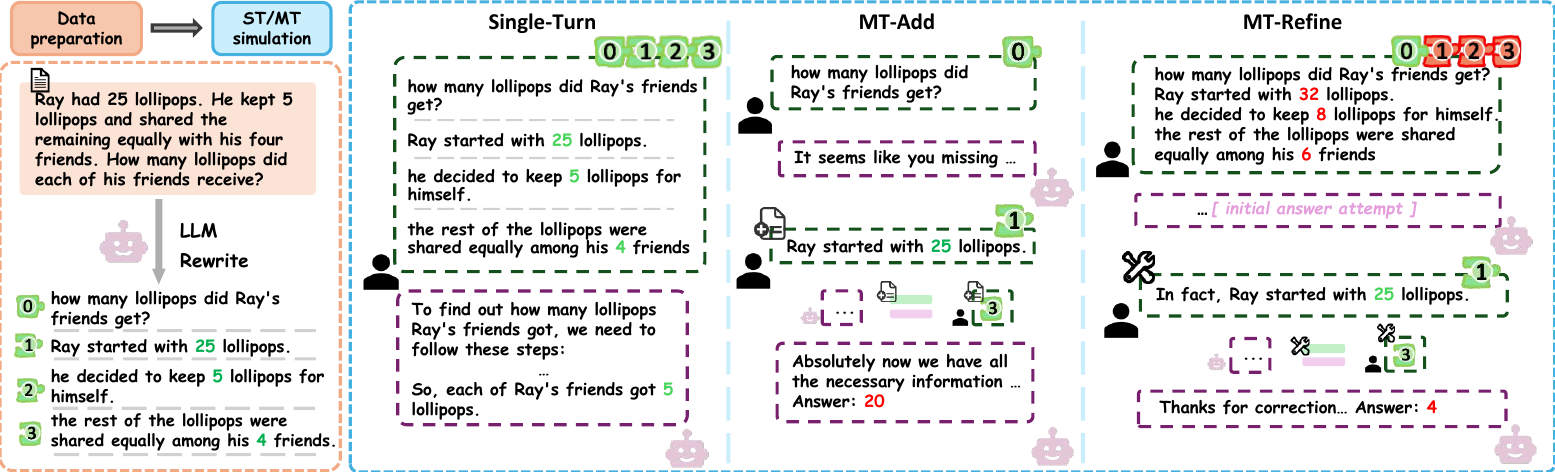}
\caption{Overview of our data preparation and multi-turn simulation pipeline. We partition single-turn prompts into segments to simulate two multi-turn scenarios: \texttt{MT-Add} (incremental information addition) and \texttt{MT-Refine} (correction of initially incorrect conditions).}
\label{fig:multiturn_setting} 
\vspace{-15pt}
\end{figure*}

\section{Related Works}

\paragraph{LLM's Vulnerability in Multi-Turn Interaction} Large Language Models (LLMs) have demonstrated significant struggles in multi-turn tasks, spanning human-model interaction \citep{laban2025llms, bai-etal-2024-mt, kwan-etal-2024-mt}, agentic inference \citep{zhou2025sweetrltrainingmultiturnllm}, and tool use \citep{patil2025the, wang2023mint}. Specifically, \citet{kwan-etal-2024-mt} introduced MT-Eval and observed an average 15\% performance degradation compared to single-turn settings. \citet{laban2025llms} further introduced a more challenging multi-turn setting involving underspecified requirements revealed sequentially, identifying a severe 39\% performance drop, they term this degradation "Lost-in-Conversation" (LiC) and attribute the root cause to premature answering attempts in early turns. In this work, we investigate both the \texttt{MT-Add} scenario, where information is added incrementally, and the \texttt{MT-Refine} scenario, where the user corrects initially incorrect conditions. Where LLM exhibits substantial vulnerability.


\paragraph{Optimization for Multi-Turn Interaction} To enhance model performance in multi-turn settings, a line of work attempts to directly implement various fine-tuning methods, including Supervised Fine-Tuning (SFT) \citep{chiang2023vicuna, ding2023enhancing, sun2024parrot}, Direct Preference Optimization \citep{shi2024directmultiturnpreferenceoptimization, xiong2024building}, and Reinforcement Learning \citep{zhou2024archer, wang2025information, kumar2024training}. While effective for general instruction following, these approaches bypass the intrinsic mechanism of degradation by relying on external guidance rather than correcting the model's internal failure modes. Moreover, methods like MT-PPO \citep{zeng2025reinforcing} and SWEET-RL \citep{zhou2025sweetrltrainingmultiturnllm} require expensive, well-designed turn-level rewards, which are unsuitable for general practical settings. Another direction explores behavioral strategies for specific scenarios, such as encouraging clarification requests \citep{wu2025collabllm} or active abstention \citep{li2025verifiable} when the information provided by users is insufficient. These methods aim to mitigate error accumulation from premature responses; however, such strategies are incompatible with scenarios requiring state updates, such as the \texttt{MT-Refine} setting addressed in this paper. In these cases, the model must generate an initial response and subsequently correct it, rather than simply remaining silent.

\section{Problem Setup}

Most existing benchmarks evaluate models using a single-turn prompt ($i^{\mathrm{single}}$), ignoring the interactive nature of practical LLM usage. In this work, we evaluate LLMs solving problems in a multi-turn setting, where users provide information $\{i_0,i_1,\ldots,i_n\}$ sequentially across turns. Specifically, we consider two scenarios: \texttt{MT-Add}, where information is added incrementally, and \texttt{MT-Refine}, where the user corrects initially incorrect conditions, as shown in Figure~\ref{fig:multiturn_setting}.

\paragraph{\texttt{MT-Add}.}
In this scenario, we partition the original single-turn prompt $i^{\mathrm{single}}$ into multiple segments $\{i_0,i_1,\ldots,i_n\}$. Here, $i_0$ represents the initial prompt specifying the problem target (e.g., ``What is Ara's total score?''), while subsequent prompts provide the additional constraints necessary to solve the problem. We assume an interaction paradigm where the user incrementally provides new information in each turn. This mirrors practical scenarios where users consistently provide follow-up information to guide the LLM toward a satisfactory answer.

\paragraph{\texttt{MT-Refine}.}
Here, we simulate error correction. The user first provides a prompt with full but corrupted information $i_0 = i^{\mathrm{corrupt}}$, where key conditions from the ground truth conditions in $\{i_1,\ldots,i_n\}$ are replaced with incorrect values. The user then corrects these specific errors sequentially through subsequent turns.

During the multi-turn conversation, information is presented sequentially. At the $k$-th turn, we use $i_k$ as the user's new query and provide the model with the conversation history $H = \{s, i_0, m_0, \ldots, i_{k-1}, m_{k-1}, i_k\}$, where $s$ is the system prompt and $m_j$ denotes the model’s response at turn $j$. We take the final-turn response $m_n$ as the final multi-turn answer. For comparison, we evaluate the model's single-turn performance by providing the full correct information $i^{\mathrm{full}}=\texttt{merge}(i_0, \ldots, i_n)$ in a single turn, taking the resulting response $m$ as the final single-turn answer.


\section{Contextual Inertia: Characterization and Quantitative Analysis}
\label{sec:multi_inv}

While recent studies have identified that LLMs struggle in multi-turn interactions \citep{laban2025llms, gupta2024llm, hankache2025evaluating}, the specific drivers of this degradation remain under-explored. Existing works primarily attribute failures to context length or premature answering \citep{laban2025llms}. In this section, we advance this understanding by analyzing the \textbf{indiscriminate nature} of the model's persistence and providing a \textbf{quantitative attribution} of these failures. We term this specific behavioral pattern as \textbf{\underline{Contextual Inertia}}: \emph{the inherent tendency of LLMs to rigidly adhere to previous reasoning traces when processing new instructions, even when those traces are partially invalid or obsolete.}

\begin{figure}[!t]
\centering
\includegraphics[width=0.98\linewidth]{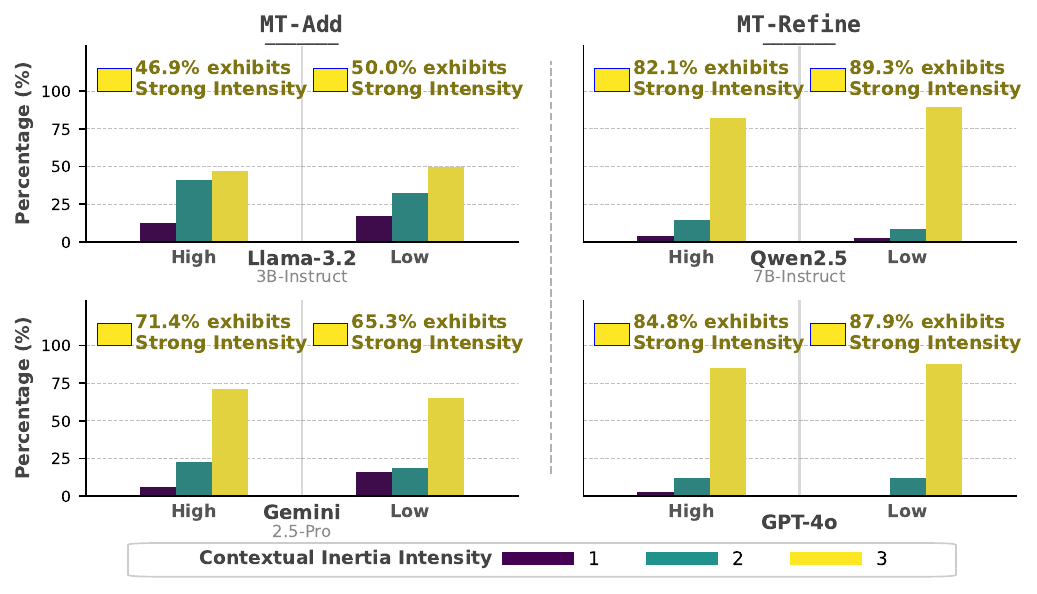}
\vspace{-7pt}
\caption{
Distributions of Contextual Inertia Intensity $\texttt{I}_{\mathrm{CI}}(m_n, m_{n-1})$. We use GPT-4o to categorize the inertia intensity as Weak (1), Moderate (2), and Strong (3). In most cases, the model's final answer $m_n$ exhibits strong inertia intensity to the preceding response $m_{n-1}$. Notably, the distribution of this intensity remains indistinguishable regardless of the conversation history quality (high vs. low), providing empirical evidence for the indiscriminate nature of contextual inertia. 
}
\label{fig:similarity} 
\vspace{-15pt}
\end{figure}

\paragraph{Indiscriminate Nature of Contextual Inertia.} 
 We first qualify and statistically analyze the contextual inertia.
Formally, let $H = \{s, i_0, m_0,\ldots, m_{n-1}, i_{n}\}$ denote a conversation history, we define the intensity of contextual inertia $\texttt{I}_{\mathrm{CI}}(m_n, m_{n-1})$ as the semantic similarity between the model's final answer $m_n$ and the preceding response $m_{n-1}$.
We categorize histories into two sets: \emph{high-quality} histories $\mathcal{H}_{\mathrm{high}}$, which result in a correct final answer $m_n \sim \pi(\cdot|H)$, and \emph{low-quality} histories $\mathcal{H}_{\mathrm{low}}$, which lead to an incorrect answer. 
We reveal the indiscriminate nature of contextual inertia, where the intensity distributions conditioned on contrasting history qualities are statistically indistinguishable. That is:
\begin{equation}
\begin{split}
&\mathbb{P}\left(\texttt{I}_{\mathrm{CI}}(m_n, m_{n-1}) \mid H \in \mathcal{H}_{\mathrm{high}}\right) \\
\approx\;\; &\mathbb{P}\left(\texttt{I}_{\mathrm{CI}}(m_n, m_{n-1}) \mid H \in \mathcal{H}_{\mathrm{low}}\right).
\end{split}
\end{equation}
To empirically validate this, we employ GPT-4o \citep{hurst2024gpt} to analyze the semantic similarity (e.g., logical structure, chain-of-thought steps) between the model's final answer $m_n$ and the preceding response $m_{n-1}$. As shown in Figure~\ref{fig:similarity}, we observe that the intensity distributions are remarkably similar across both groups. Moreover, the model's final answer $m_n$ and the last response $m_{n-1}$ exhibit high similarity in most cases. This validates the indiscriminate nature of contextual inertia: even when $m_{n-1}$ is misleading or erroneous, the model is compelled to propagate these traces into subsequent turns.

\paragraph{Quantifying the Dominance of Contextual Inertia.}
Beyond statistical similarity, we further trace the sources of error by collecting multi-turn conversations $\{s, i_0,\dots, i_{n}, m_n\}$ where the final answer $m_n$ is verified as incorrect, and prompt GPT-4o to classify the root cause into three categories:\footnote{To isolate the limitations of the base model, we select conversations where providing full information $i^{\mathrm{full}} = \mathrm{merge}(i_0, \dots, i_n)$ in a single turn yields high (> 0.7) accuracy.}


\begin{itemize}[nosep, leftmargin=10pt] 
    \item \textbf{Misleading Context:} The failure stems from previous responses $\{m_0,\dots,m_{n-1}\}$ which, while ostensibly correct, create a misleading or contradictory context for follow-up information.
    \item \textbf{Propagated Error:} The failure is caused by previous responses that are fundamentally wrong, where the model passively inherits and propagates these factual or logical errors to the final response.
    \item \textbf{Local Reasoning Failure:} Previous responses are of generally good quality; the failure occurs independently in the final turn when the model processes the user's latest query $i_{n}$ (e.g., losing the problem target or making arithmetic errors).
\end{itemize}

\begin{figure}[!t]
\centering
\includegraphics[width=\linewidth]{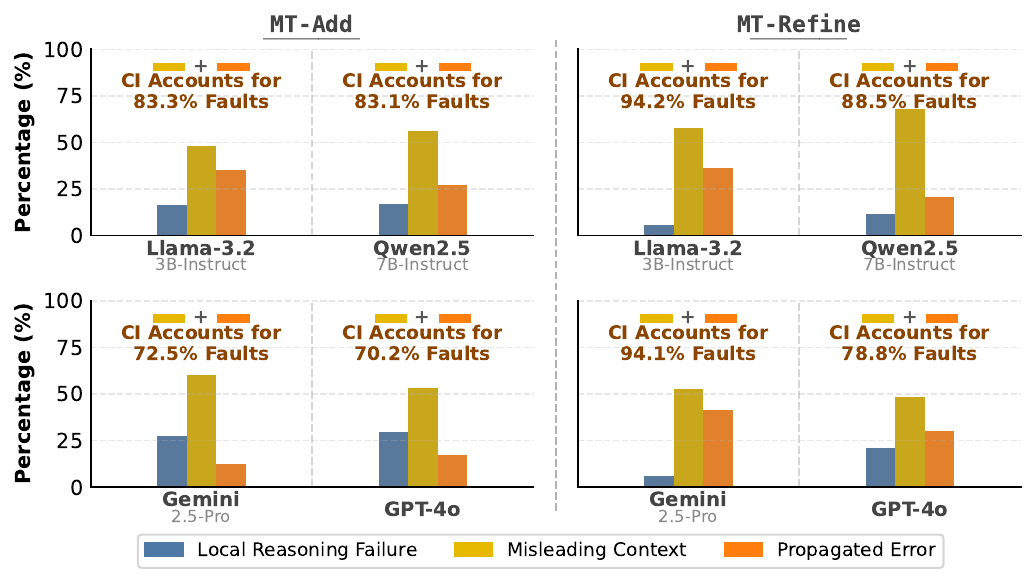}
\vspace{-20pt}
\caption{Root cause of failures in multi-turn conversations: failures predominantly originate from previous responses (\textbf{Misleading Context} and \textbf{Propagated Error}), which are driven by Contextual Inertia.}
\label{fig:root_cause_failures} 
\vspace{-15pt}
\end{figure}

As demonstrated in Figure~\ref{fig:root_cause_failures}, we find that over 70\%--90\% of multi-turn errors can be directly traced back to previous responses (\textbf{Misleading Context} or \textbf{Propagated Error}). In these cases, due to the indiscriminate nature, contextual inertia compels the LLM to adopt misleading or erroneous traces without verification, causing error accumulation across the conversation that ultimately degrades performance.

Due to its indiscriminate nature, Contextual Inertia serves as the primary driver of model vulnerability in multi-turn settings. While recent studies \citep{wanmitigating,laban2025llms,castillo-bolado2024beyond} demonstrate that LLMs tend to attend heavily to the last response, we go a step further by revealing the indiscriminate nature of this phenomenon in multi-turn conversations. This insight motivates our approach to break this inertia for stable interaction. We detail our methodology in Section~\ref{sec:method}.



\section{Approach}
\label{sec:method}

\begin{figure}[t]
\centering
\includegraphics[width=\linewidth]{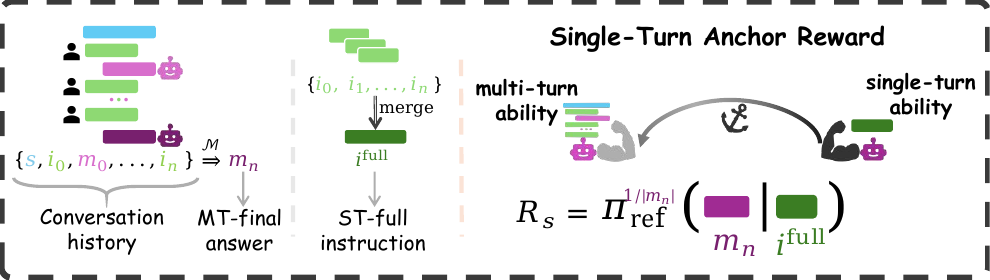}
\vspace{-15pt}
\caption{\textbf{Single-Turn Anchor Reward ($R_s$).} We leverage the model's superior single-turn ability on full instruction ($i^{\mathrm{full}}$) as anchor for the multi-turn final answer $m_n$. Our filtering strategy (Equation~\ref{eq:data_filter}) ensures the anchor is reliable by retaining only histories where single-turn performance exceeds multi-turn.}
\label{fig:sta} 
\vspace{-10pt}
\end{figure}

To address \textit{Contextual Inertia} in multi-turn LLMs, where models rigidly adhere to obsolete reasoning traces, we propose \textbf{R}einforcement \textbf{L}earning with \textbf{S}ingle-\textbf{T}urn \textbf{A}nchors (\textbf{RLSTA}).

Existing solutions typically attribute multi-turn vulnerability to ``uncertainty'', mitigating it through abstention \citep{li2025verifiable} or clarification requests \citep{wu2025collabllm} in early turns to avoid generating premature answers based on insufficient information. While viable for handling ambiguity in \texttt{MT-Add}, these strategies merely address the symptoms rather than curing the root cause: Contextual Inertia. Furthermore, such approaches are incompatible with scenarios like \texttt{MT-Refine}, where the interaction relies on the model providing an initial response that the user subsequently corrects (e.g., \textit{``No, actually X is 5''}). In these cases, the model cannot abstain in early turns; it must fundamentally overcome its inertia for a correct answer.


We target this root cause directly. Instead of resorting to passive abstention, we aim to \textit{calibrate} the model's multi-turn reasoning against its own superior single-turn capabilities. Our approach combines: (i) Latent Capability Filtering, which isolates cases where the model successfully solves the problem with the full instruction ($i^{\mathrm{full}}$), yet fails to generate the correct answer under the sequential multi-turn history ($H$) due to contextual inertia. (ii) An anchor-based RL method that utilizes the model's single-turn reasoning ability as internal anchors. We detail our approach as follows.

\subsection{Latent Capability Filtering}
In Section~\ref{sec:multi_inv}, we identify contextual inertia within potentially misleading or erroneous conversation histories as the primary driver of multi-turn vulnerability in LLMs. Building on this insight, we aim to rectify such behavior by specifically optimizing the last-turn response $m_n$ in cases where the multi-turn conversation history $H = \{s, i_0, m_0, \ldots, i_{n-1}, m_{n-1}, i_n\}$ leads to performance degradation.


We first collect a raw set of multi-turn conversation histories $\mathcal D_{\mathrm{raw}}$. 
We then perform \textbf{Latent Capability Filtering} to ensure we can use the model's single-turn ability as a stable anchor for the model's final response given multi-turn conversation history. Specifically, we isolate instances where the model possesses the latent capability to solve the problem given the full information $i^{\mathrm{full}} = \mathtt{merge}(i_0,\cdots,i_{n-1},i_n)$, yet fails to do so under the original multi-turn history $H$. Formally, we retain conversation histories satisfying the following condition:
\begin{equation}
    \mathbb E_{\scriptscriptstyle m \sim \pi(\cdot| i^{\mathrm{full}})} [{\scriptstyle\mathtt{Ver}}(m)] > \mathbb E_{\scriptscriptstyle m_n \sim \pi(\cdot| H)}[ {\scriptstyle\mathtt{Ver}}(m_n)].
    \label{eq:data_filter}
\end{equation}
Where $\mathtt{Ver}(\cdot)$ is the verifier to check the correctness of the final answer with $0/1$, this filtering process yields the dataset $\mathcal D_{\mathcal M}$ for model $\mathcal M$.
Our objective is to break contextual inertia in scenarios where the preceding conversation is low quality. Therefore, we exclude conversation histories $H$ where single-turn and multi-turn performances are comparable. This exclusion allows us to focus on instances where the model possesses superior single-turn capabilities, thereby providing a high-quality supervision signal for alignment, which we formulate as our single-turn anchor for RL training.

\subsection{RL with Single-Turn Anchors}

We use GRPO as our training algorithm. For a given conversation history $H$, GRPO samples a group of outputs $\{m_n^{(1)}, \ldots,m_n^{(G)}\}$ from the old policy $\theta_{\mathrm{old}}$ and optimizes the policy model by maximizing:
\begin{equation}
\begin{split}
    &{\scriptstyle\mathcal J_{\mathrm{GRPO}}(\theta)=\scriptstyle\mathbb E\left[H \sim \mathcal D_{\mathcal M}, \{m_n^{(i)}\}_{i=1}^G \sim \pi_{\theta_{\mathrm{old}}}(\cdot|H)\right] } \\
    &{\scriptstyle \sum_{i = 1}^G\limits \frac{1}{|m_n^{(i)}|}  \sum_{t = 1}^{|m_n^{(i)}|}\limits} {\{}{\scriptstyle \min\left[r_{i,t}(\theta) \hat{A}_{i,t}, \:\:\mathrm{clip}\left( r_{i,t}(\theta), 1-\epsilon, 1+ \epsilon \right) \hat{A}_{i,t}\right]} \\
    &{\scriptstyle- \beta \mathbb D_{\mathrm{KL}}[\pi_\theta||\pi_{\mathrm{ref}}]} {\}},
\end{split}
\end{equation}
where
\begin{equation*}
{\scriptstyle r_{i,t}(\theta) = \frac{\pi_\theta \left(m_{n,t}^{(i)} | H, m_{n,<t}^{(i)}\right)}{\pi_{\theta_{\mathrm{old}}} \left(m_{n,t}^{(i)} | H, m_{n,<t}^{(i)}\right)}, \quad \hat{A}_{i,t} = \frac{R_i - \mathrm{mean}\left(\{R_i\}_{i=1}^G\right)}{\mathrm{std}\left(\{R_i\}_{i=1}^G\right)}}.
\end{equation*}
Here, $R_i$ denotes the reward for response $m_{n}^{(i)}$, which is typically a rule-based reward, such as using a verifier to check the final answer or to check whether the response follows a required format \citep{guo2025deepseek, zeng2025simplerlzoo}.

In addition to the outcome-based verification reward $R_{\mathrm{v}} = \mathtt{Ver}(m_{n}^{(i)})$, we introduce a mechanism to stabilize the reasoning process: the \textbf{Single-Turn Anchor Reward} ($R_{\mathrm{s}}$). This component utilizes the model's strong ability under the full-information single-turn setting ($i^{\mathrm{full}}$) as a stable anchor for the current multi-turn generation $m_n^{(i)}$:
\begin{equation}
\small R_{\mathrm{s}} = \left( \prod_{t = 1}^{|m_n^{(i)}|} \pi_{\theta_{\mathrm{ref}}} \left(m_{n,t}^{(i)} \mid i^{\mathrm{full}}, m_{n,<t}^{(i)}\right) \right)^{\frac{1}{|m_n^{(i)}|}},
\label{eq:r_single}
\end{equation}
Here, $i^{\mathrm{full}}$ merges all user conditions from the multi-turn history $H$ into a single-turn query, and $\pi_{\theta_{\mathrm{ref}}}$ denotes the base model. By calculating the length-normalized likelihood of the response under $\pi_{\theta_{\mathrm{ref}}}(\cdot \mid i^{\mathrm{full}})$, $R_{\mathrm{s}}$ quantifies how well the multi-turn response aligns with the model's superior single-turn capabilities.

Combining the outcome reward $R_{\mathrm{v}}$ and the anchor reward $R_{\mathrm{s}}$ yields the final reward for response $m_{n}^{(i)}$:
\begin{equation}
R = R_{\mathrm{v}} + \alpha R_{\mathrm{s}},
\end{equation}
where $\alpha$ is a hyperparameter. We set $\alpha = 1/G$ so that the advantage $\hat{A}$ for a correct response with $R_{\mathrm{v}} = 1$ remains positive.

We highlight that $R_{\mathrm{s}}$ serves as a critical behavioral \textbf{anchor}. By leveraging our data-filtering strategy (Equation~\ref{eq:data_filter}), which ensures that the single-turn policy $\pi_{\theta_{\mathrm{ref}}}(\cdot \mid i^{\mathrm{full}})$ demonstrates superior accuracy compared to the noisy multi-turn history $\pi_{\theta}(\cdot \mid H)$, $R_{\mathrm{s}}$ provides a robust supervision signal. This signal effectively pulls the generation process away from the bias of contextual inertia, anchoring the model to the correct reasoning path established in the single-turn setting.

\section{Experiments}
\label{sec:exp}
\begin{table*}[t]
\centering
\caption{Experimental results on \texttt{MT-Add} and \texttt{MT-Refine} Tasks. We compare RLSTA with SFT, DPO, and vanilla GRPO. \textbf{Bold} indicates best performance, \underline{underline} indicates improvement over Base, and \textcolor{gray}{gray} indicates performance lower than Base.}
\vspace{-10pt}
\resizebox{\textwidth}{!}{%
\begin{tabular}{lcccccccccccccccc}
\toprule

\multirow{3}{*}{\textbf{Method}} & 
\multicolumn{4}{c}{\textbf{Qwen2.5-3B-Instruct}} & 
\multicolumn{4}{c}{\textbf{Qwen2.5-7B-Instruct}} & 
\multicolumn{4}{c}{\textbf{Qwen3-4B-Instruct-2507}} & 
\multicolumn{4}{c}{\textbf{Llama-3.2-3B-Instruct}} \\
\cmidrule(lr){2-5} \cmidrule(lr){6-9} \cmidrule(lr){10-13} \cmidrule(lr){14-17}
\addlinespace[-2.5pt]
 & \multicolumn{2}{c}{\small \textbf{\texttt{MT-Add}}} & \small \textbf{\texttt{MT-Refine}} & \multirow{2}{*}{ \textbf{Avg}} 
 & \multicolumn{2}{c}{\small \textbf{\texttt{MT-Add}}} & \small \textbf{\texttt{MT-Refine}} & \multirow{2}{*}{ \textbf{Avg}} 
 & \multicolumn{2}{c}{\small \textbf{\texttt{MT-Add}}} & \small \textbf{\texttt{MT-Refine}} & \multirow{2}{*}{ \textbf{Avg}} 
 & \multicolumn{2}{c}{\small \textbf{\texttt{MT-Add}}} & \small \textbf{\texttt{MT-Refine}} & \multirow{2}{*}{ \textbf{Avg}} \\
\cmidrule(lr){2-3} \cmidrule(lr){4-4} 
\cmidrule(lr){6-7} \cmidrule(lr){8-8} 
\cmidrule(lr){10-11} \cmidrule(lr){12-12} 
\cmidrule(lr){14-15} \cmidrule(lr){16-16}
\addlinespace[-2.5pt]
 & \small Math & \small Code & \small Math & 
 & \small Math & \small Code & \small Math & 
 & \small Math & \small Code & \small Math & 
 & \small Math & \small Code & \small Math & \\
\midrule

Base & 0.493 & 0.220 & 0.603 & 0.439 & 0.638 & 0.312 & 0.669 & 0.540 & 0.772 & 0.370 & 0.814 & 0.652 & 0.470 & 0.140 & 0.585 & 0.398 \\
SFT  & \underline{0.546} & \textcolor{gray}{0.203} & \textcolor{gray}{0.533} & 0.427 & \underline{0.740} & \textcolor{gray}{0.310} & \underline{0.694} & 0.581 & \underline{0.826} & \underline{0.444} & \textcolor{gray}{0.809} & 0.693 & \underline{0.568} & \underline{0.177} & \textcolor{gray}{0.519} & 0.421 \\
DPO  & \underline{0.546} & \textcolor{gray}{0.203} & \textcolor{gray}{0.568} & 0.439 & \textcolor{gray}{0.633} & 0.312 & \textcolor{gray}{0.522} & 0.489 & \underline{0.784} & \underline{0.416} & \textcolor{gray}{0.801} & 0.667 & \underline{0.559} & \underline{0.198} & \textcolor{gray}{0.368} & 0.375 \\
GRPO & \underline{0.636} & \textcolor{gray}{0.190} & \underline{0.734} & 0.520 & \underline{0.803} & \underline{0.336} & \textbf{0.836} & 0.659 & \underline{0.839} & \underline{0.472} & \underline{0.882} & 0.731 & \underline{0.608} & \underline{0.190} & \underline{0.620} & 0.473 \\
\midrule
\textbf{RLSTA (Ours)} & \textbf{0.715} & \textbf{0.256} & \textbf{0.745} & \textbf{0.572} & \textbf{0.857} & \textbf{0.350} & \underline{0.822} & \textbf{0.676} & \textbf{0.903} & \textbf{0.552} & \textbf{0.898} & \textbf{0.784} & \textbf{0.649} & \textbf{0.205} & \textbf{0.640} & \textbf{0.498} \\
\bottomrule
\end{tabular}%
}
\vspace{-10pt}

\label{tab:main_results}

\end{table*}
We evaluate our method across various multi-turn scenarios, tasks, and training configurations in controlled settings. The results demonstrate that our method effectively aligns the model's multi-turn performance, yielding superior results and promising generalization to other tasks, while preserving the base model's single-turn performance and long-context capabilities. We detail our experimental setup in Subsection~\ref{sec:exp_setup}, followed by the main results and comparisons with existing training methods and other models in Subsection~\ref{sec:main_results}. Finally, we examine the preservation of long-context capabilities and analyze the training dynamics of the single-turn anchor reward in Subsection~\ref{sec:efficiency}.

\subsection{Experiments Setup}
\label{sec:exp_setup}

Here we introduce the tasks, datasets, and models used in our experiments. Additional details are provided in Appendix~\ref{sec:add_exp_details}.

\paragraph{Data Preparation.}
We construct our dataset following the instruction segmentation protocol proposed by \citet{laban2025llms}. For training, we derive multi-turn samples from the GSM8K dataset \citep{cobbe2021training} by employing GPT-4o \citep{hurst2024gpt} to decompose original single-turn queries into sequential instructions. For evaluation, we adopt the multi-turn benchmarks from \citet{laban2025llms}. Notably, while our model is trained exclusively on math-domain multi-turn scenarios, we evaluate it on diverse domains, including code and summarization, to assess its cross-domain generalization capabilities.

\paragraph{Multi-turn Scenario Simulation.}
We simulate two distinct interaction patterns: \texttt{MT-Add} and \texttt{MT-Refine}. In the \texttt{MT-Add} setting, segmented instructions $\{i_0,i_1,\cdots,i_n\}$ are presented sequentially, requiring the model to integrate new constraints incrementally. The \texttt{MT-Refine} setting simulates error correction. Here, we employ GPT-4o to alter key conditions within the segments $\{i_1,\cdots,i_n\}$ and concatenate these modified conditions with the initial prompt $i_0$. This forms a corrupted initial instruction $i^{\mathrm{corrupt}}$, while the original valid segments are provided in subsequent turns as user corrections. Note that the code and summary tasks are evaluated solely under the \texttt{MT-Add} setting due to data compatibility. Beyond fixed datasets, following \citet{laban2025llms}, we also assess performance in a dynamic setting using GPT-4o-mini as a user simulator. This simulator interactively selects instructions rather than following a fixed sequence, thereby more closely mimicking realistic human interactions.

\paragraph{Models and Baselines.}
We validate our method across a range of open-weights models, including Qwen2.5-3B/7B-Instruct \citep{Yang2024Qwen25TR}, Qwen3-4B-2507 \citep{yang2025qwen3}, and Llama-3.2-3B-Instruct \citep{Dubey2024TheL3}. We employ GRPO as our primary training algorithm and compare RLSTA against standard fine-tuning baselines: Supervised Fine-Tuning (SFT), Direct Preference Optimization (DPO) \citep{rafailov2023direct}, and vanilla GRPO. For SFT and DPO, we construct training pairs by appending the correct single-turn response to the multi-turn history, and utilize the model's original erroneous response as the negative sample for DPO. Additionally, we benchmark against uncertainty-handling strategies: RLAAR \citep{li2025verifiable} and CollabLLM \citep{wu2025collabllm}.

\subsection{Main Results}
\label{sec:main_results}

\begin{figure}[t]
\centering
\includegraphics[width=\linewidth]{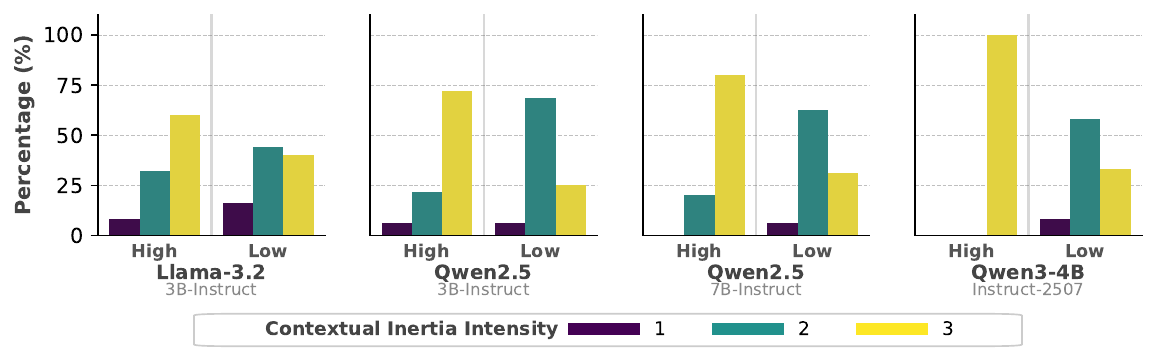}
\vspace{-20pt}
\caption{The distributions of contextual inertia intensity after RLSTA. Compared with base model in Figure~\ref{fig:similarity}, RLSTA successfully breaks the contextual inertia, while preserving the ability to utilize beneficial history.}
\label{fig:Similarity_RLSTA} 
\vspace{-15pt}
\end{figure}

\paragraph{RLSTA Can Effectively Break Contextual Inertia.}
We begin by analyzing the distributions of contextual inertia intensity after RLSTA on \texttt{MT-Add}. Following the methodology in Section~\ref{sec:multi_inv}, we categorize conversation histories based on the base model's performance: ``High'' quality histories (leading to correct answers) and ``Low'' quality ones (leading to incorrect answers). While the base model exhibits indiscriminate inertia regardless of the quality of conversation history (Figure~\ref{fig:similarity}), RLSTA maintains high similarity for the High-quality group but shows notably lower contextual inertia intensity for the Low-quality group (Figure~\ref{fig:Similarity_RLSTA}). This confirms that RLSTA successfully break the indiscriminate nature of contextual inertia, while preserving the ability to utilize beneficial history.

\paragraph{RLSTA Outperforms Standard Fine-Tuning Methods.} As shown in Table~\ref{tab:main_results}, RLSTA significantly improves the model's performance across different multi-turn scenarios compared to SFT, DPO, and GRPO. Notably, although we only train the model using math-domain multi-turn scenarios, RLSTA demonstrates strong generalization to other domains (e.g., Code). This suggests that RLSTA fosters a fundamental resilience against contextual inertia on misleading conversation history, enabling the model to maintain robust reasoning capabilities even in domains outside its training distribution.

\begin{table}[h]
    \centering
    \caption{Comparison of LiC Scores ($\uparrow$) between RLSTA and RLAAR on \texttt{MT-Add} scenario. \underline{Underline} indicates improvement over the Base Model.}
        
    \vspace{-5pt}
    \resizebox{\columnwidth}{!}{
        \begin{tabular}{lccccc}
            \toprule
            \textbf{Method} & \textbf{Math} & \textbf{Actions} & \textbf{Database} & \textbf{Code} & \textbf{Avg} \\
            \midrule
            Base & 0.728 & 0.365 & 0.564 & 0.525 & 0.546 \\
            RLAAR &  \underline{0.950} &  \underline{0.530} & \underline{0.873} & \underline{0.702} &  \underline{0.764} \\
            RLSTA & \underline{1.001} & \underline{0.688} &  \underline{0.691} &  \underline{0.596} &  \underline{0.744} \\
            RLSTA (+ Abstain) & \underline{0.988} & \underline{0.702} &  \underline{0.762} &  \underline{0.642} &  \underline{0.773} \\
            \bottomrule
        \end{tabular}
    }
    \vspace{-15pt}
    
    \label{tab:rlaar_comparison}
\end{table}

\begin{table}[h]
    \centering    
    \vspace{-5pt}
    \caption{Comparison between RLSTA and CollabLLM on \texttt{MT-Add} scenario, we report model performance ($\uparrow$) and token consumption ($\downarrow$)  (\perf{Performance}{Tokens})}
    \vspace{-5pt}
    \resizebox{\columnwidth}{!}{
        \setlength{\tabcolsep}{12pt} 
        \begin{tabular}{ccccc} 
            \toprule
            \multirow{2}{*}{\textbf{Method}} & \multicolumn{2}{c}{\textbf{Code}} & \multicolumn{2}{c}{\textbf{Math}} \\
            \cmidrule(lr){2-3} \cmidrule(lr){4-5}
            & \textbf{Sing-turn} & \textbf{Multi-turn} & \textbf{Sing-turn} & \textbf{Multi-turn} \\
            \midrule
            Base      & \perf{0.320}{142.3} & \perf{0.223}{360.2} & \perf{0.782}{224.8} & \perf{0.443}{319.9} \\ \addlinespace[0.5em]
            CollabLLM & \perf{0.302}{125.2} & \perf{0.214}{\textbf{321.5}} & \perf{0.758}{250.0} & \perf{0.433}{\textbf{294.6}} \\ \addlinespace[0.5em]
            RLSTA      & \perf{0.343}{149.4} & \perf{\textbf{0.309}}{339.3} & \perf{0.754}{230.4} & \perf{\textbf{0.654}}{374.0} \\
            \bottomrule
        \end{tabular}
    }
    \vspace{-10pt}

    \label{tab:collabllm_results}
\end{table}

\paragraph{RLSTA Achieves Comparable Performance to Abstention and Active Inquiry Methods.}
We further benchmark our method against uncertainty-handling strategies: RLAAR \citep{li2025verifiable}, which encourages abstention under uncertainty, and CollabLLM \citep{wu2025collabllm}, which tunes the model to actively request more information. 
For RLAAR, we reference the results reported in \citep{li2025verifiable} using Qwen2.5-7B-Instruct as the base model. As presented in Table~\ref{tab:rlaar_comparison}, RLSTA achieves an average LiC Score (the ratio between multi and single turn performance) comparable to RLAAR, while significantly outperforming it on the \textit{Math} and \textit{Actions} tasks. 
Although RLAAR maintains a marginal advantage in Code and Database by leveraging code training data, RLSTA achieves competitive results without such domain-specific supervision, demonstrating strong cross-domain generalization capabilities.
Furthermore, when explicitly instructing the model to abstain via the system prompt (RLSTA + Abstain), our method yields higher average performance than RLAAR.
Regarding CollabLLM, we employ an LLM-based user simulator to create realistic multi-turn scenarios, evaluating both performance and token efficiency using Llama-3.1-8B-Instruct as the base model. As detailed in Table~\ref{tab:collabllm_results}, although CollabLLM reduces token consumption, it fails to deliver performance improvements on our \texttt{MT-Add} tasks under this dynamic setting. In contrast, RLSTA achieves substantial gains (36.9\% on code, 47.6\% on math) with competitive efficiency (utilizing 5.8\% fewer tokens for code with only a moderate 17.2\% increase for math). Crucially, unlike these baselines, RLSTA improves multi-turn performance without relying on passive abstention, thereby serving as a generalizable approach to a wider range of scenarios, like \texttt{MT-Refine}, to which uncertainty-handling strategies are inapplicable.

\subsection{Long-Context Preservation and Training Efficiency}
\label{sec:efficiency}

\begin{table}[h]
\centering
\vspace{-10pt}
\caption{Model performance (Coverage Score $\uparrow$) on the Summary Task across multi and single turn scenarios. RLSTA maintains, and in most cases improves, the model's long-context capabilities while simultaneously stabilizing multi-turn interactions.}
\vspace{-7pt}
\resizebox{0.95\columnwidth}{!}{
\begin{tabular}{ll ccc}
\toprule
\textbf{Model} & \textbf{Method} & \textbf{Multi-turn} & \textbf{Single-turn} & \textbf{LiC Score} \\
\midrule
\multirow{2}{*}{\begin{tabular}[l]{@{}l@{}} Qwen2.5-3B \\[-0.6ex] {\color{gray} Instruct}\end{tabular}} 
    & { Base} & 0.359 & 0.524 & 0.685 \\
    & RLSTA & 0.433 & 0.520 & \textbf{0.832} \\
\midrule
\multirow{2}{*}{\begin{tabular}[l]{@{}l@{}} Qwen2.5-7B \\[-0.6ex] {\color{gray} Instruct}\end{tabular}} 
    & Base & 0.446 & 0.594 & 0.751 \\
    & RLSTA & 0.441 & 0.590 & 0.747 \\
\midrule
\multirow{2}{*}{\begin{tabular}[l]{@{}l@{}}Qwen3-4B \\[-0.6ex] {\color{gray}Instruct-2507}\end{tabular}} 
    & Base & 0.555 & 0.750 & 0.740 \\
    & RLSTA & 0.632 & 0.758 & \textbf{0.835} \\
\midrule
\multirow{2}{*}{\begin{tabular}[l]{@{}l@{}} Llama-3.2-3B \\[-0.6ex] {\color{gray} Instruct}\end{tabular}} 
    & Base & 0.384 & 0.549 & 0.698 \\
    & RLSTA & 0.434 & 0.555 & \textbf{0.782} \\
\bottomrule
\end{tabular}
}
\vspace{-10pt}

\label{tab:summary_results}
\end{table}
\paragraph{RLSTA Maintains Long-Context Processing Ability.} We further check whether breaking contextual inertia on potentially erroneous history affects the model's ability to process long contexts, we evaluate performance on the multi-turn Summary task. Here, the model is provided with very long contexts (spanning tens of thousands of tokens) across multiple turns and instructed to summarize the previous context in each turn. We use \emph{Coverage Scores} \citep{laban2024summary} as the metric, which evaluates the model's capacity to process long contexts. As shown in Table~\ref{tab:summary_results}, compared to the Base model, RLSTA maintains or even improves model's long-context processing ability. This demonstrates that RLSTA successfully breaks contextual inertia without compromising the model's ability to utilize conversation history in long-context scenarios.

\begin{figure}[ht]
\centering
\includegraphics[width=\linewidth]{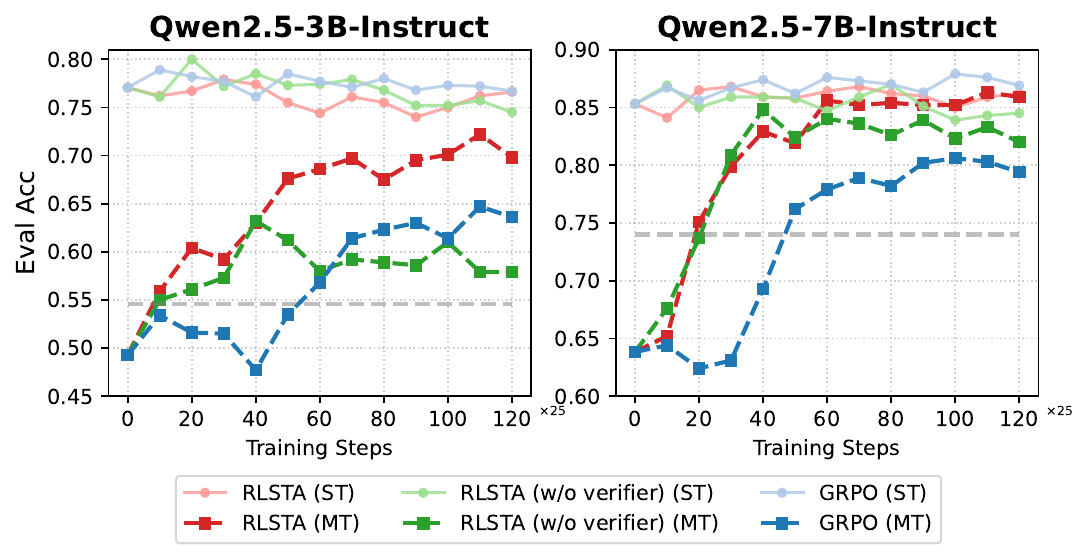}
\vspace{-23pt}
\caption{Training dynamics of Single-turn (ST) and Multi-turn (MT) performance across different methods. The gray dashed line (\protect\tikz[baseline=-0.5ex]\protect\draw[gray,dashed,line width=1pt] (0,0) -- (0.55,0);) represents the corresponding SFT MT performance baseline. Notably, even in the absence of external verifiable rewards (\textbf{RLSTA w/o verifier}), our method effectively narrows the performance gap between single-turn and multi-turn settings.}
\label{fig:no_verifier} 
\vspace{-15pt}
\end{figure}

\paragraph{Training Dynamics Comparison.} Finally, we examine the training dynamics and the efficiency of our single-turn anchor reward $R_s$. Figure~\ref{fig:no_verifier} illustrates the trajectory of Single-Turn (ST) and Multi-Turn (MT) performance during training. We observe that compared to standard GRPO, RLSTA exhibits accelerated convergence and superior asymptotic performance.

Crucially, we evaluate a variant, \textbf{RLSTA (w/o verifier)}, which relies solely on the internal single-turn anchor reward ($R_s$) without access to the external ground-truth verifier ($R_v$).The results demonstrate that even in the absence of external verifiable rewards, RLSTA effectively stabilizes multi-turn interaction and even achieves performance comparable to the full RLSTA setting, highlighting the potential of RLSTA in general domains where external verifiers are unavailable.


\section{Conclusion}
In this work, we identified the indiscriminate nature of \emph{Contextual Inertia} and quantitatively attributed its impact as the primary cause of LLM vulnerability in multi-turn interaction. We proposed Reinforcement Learning with Single-Turn Anchors to address this issue by leveraging the intrinsic single-turn capabilities of the model as stable internal anchors. Extensive experiments demonstrate that our approach significantly enhances stability in both \texttt{MT-Add} and \texttt{MT-Refine} scenarios while outperforming standard fine-tuning baselines. Moreover, our method exhibits strong cross-domain generalization and data efficiency even in the absence of external verifiers. These findings highlight the potential of our framework to facilitate more reliable and adaptive multi-turn interactions for general applications.


\section*{Limitations}

The effectiveness of RLSTA is intrinsically bounded by the model's single-turn capabilities, as our method leverages the single-turn response as a supervisory anchor and thus assumes the model possesses the latent knowledge to solve the problem given full information. Additionally, our current framework operates under a passive interaction paradigm where the user voluntarily provides all necessary conditions, leaving scenarios that require proactive clarification unaddressed. Furthermore, while RLSTA offers a universal approach to breaking contextual inertia across various settings like \texttt{MT-Add} and \texttt{MT-Refine}, it does not yet incorporate meta-cognitive decision-making; given that abstention-based methods can be effective in specific multi-turn scenarios (\texttt{MT-Add}), teaching the model to dynamically evaluate context sufficiency and select the optimal strategy, such as reasoning, clarifying, or abstaining, remains a promising direction for future investigation.



\bibliography{reference}

\appendix


\section{Additional Experiments}




In this section, we present additional experiments to comprehensively investigate the nature of Contextual Inertia and validate the robustness of RLSTA. We begin by \textbf{quantifying the intrinsic vulnerability} of LLMs in multi-turn settings. We then extend our analysis of Contextual Inertia to include \textbf{more advanced models}. To further validate our approach, we examine the efficacy of RLSTA \textbf{without data filtering} and its generalization to reasoning-oriented \textbf{``Thinking'' models}. Furthermore, we investigate the impact of our method on \textbf{single-turn capabilities} and corroborate our observations using \textbf{alternative evaluators} (gemini-2.5-pro \citep{comanici2025gemini}). Finally, we demonstrate the robustness of our filtering strategy through its \textbf{transferability across heterogeneous data sources}.

\subsection{Vulnerability of LLMs in multi-turn settings}
\begin{figure}[h]
\centering
\vspace{-15pt}
\includegraphics[width=\linewidth]{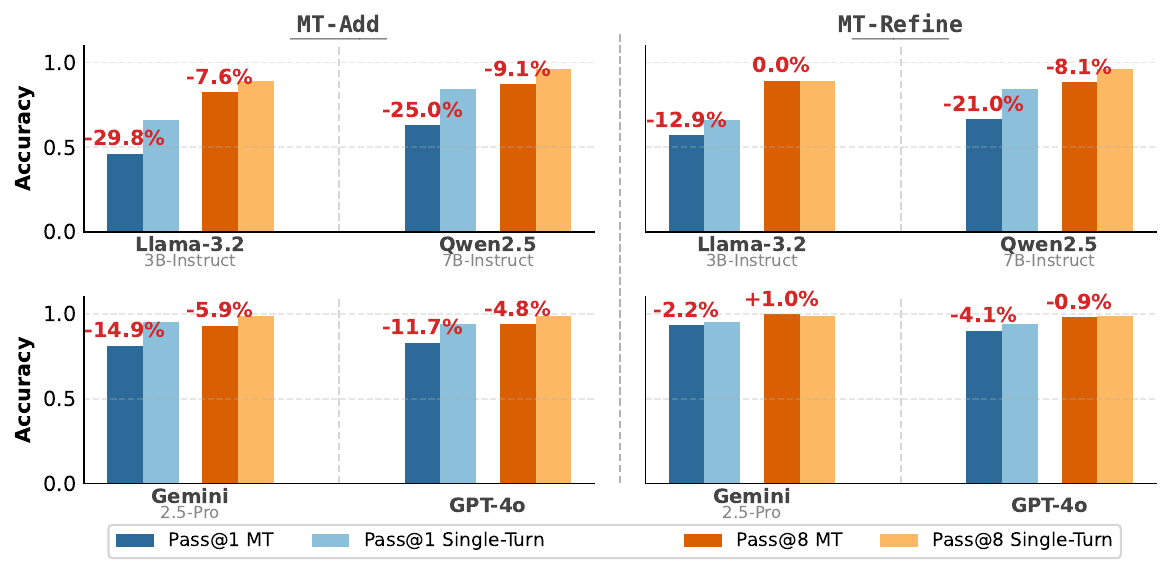}
\vspace{-20pt}
\caption{Vulnerability of LLMs in multi-turn settings: While Pass@1 drops significantly compared to single-turn, Pass@8 remains relatively stable. This indicates that the model retains the latent capability to solve the problem but is biased by the decoding path.}
\label{fig:mt_performance} 
\vspace{-5pt}
\end{figure}

We begin by quantifying the vulnerability of LLMs in multi-turn settings. As illustrated in Figure~\ref{fig:mt_performance}, model performance exhibits substantial degradation in multi-turn settings compared to single-turn baselines. Specifically, the pass@1 performance suffers a precipitous drop, with an average decline of over 20\% in the \texttt{MT-Add} and 10\% in the \texttt{MT-Refine}. However, this degradation is largely mitigated when sampling is permitted: the pass@8 performance exhibits a much milder drop of less than 7\% in \texttt{MT-Add} and merely 2\% in \texttt{MT-Refine}. This discrepancy reveals a critical insight: the model retains the latent capability to solve the problem (evidenced by the preserved pass@8 score), but its most probable decoding trajectory is effectively ``hijacked'' by the conversation history.

\subsection{Contextual Inertia Analysis}

\begin{figure}[h]
\centering
\includegraphics[width=\linewidth]{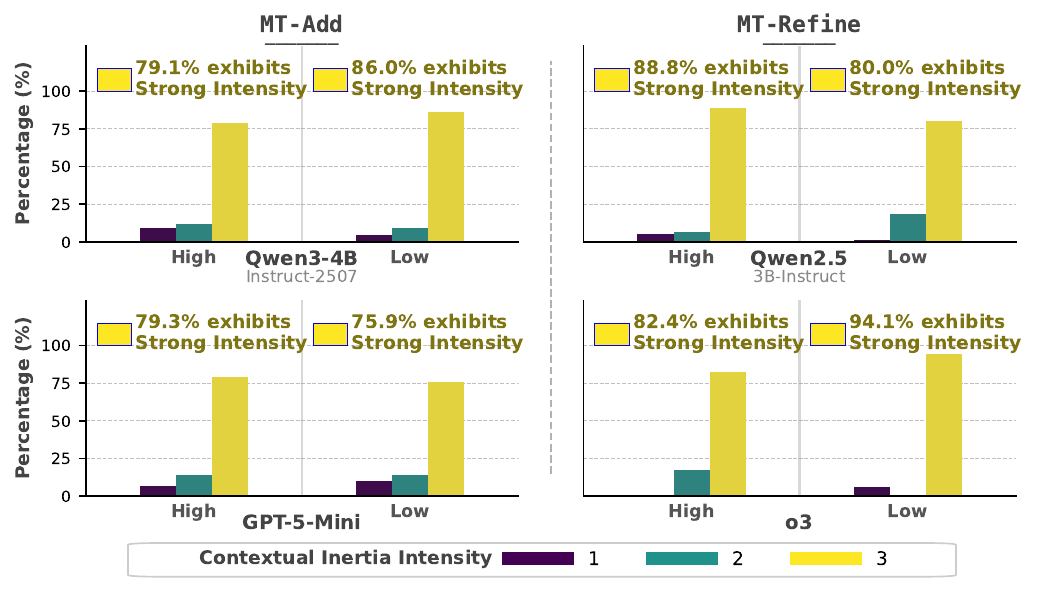}
\caption{Additional experiments: the distribution of the contextual inertia intensity for more models.}
\label{fig:Similarity_Vulnerability_appdix} 
\end{figure}

\begin{figure}[h]
\centering
\includegraphics[width=\linewidth]{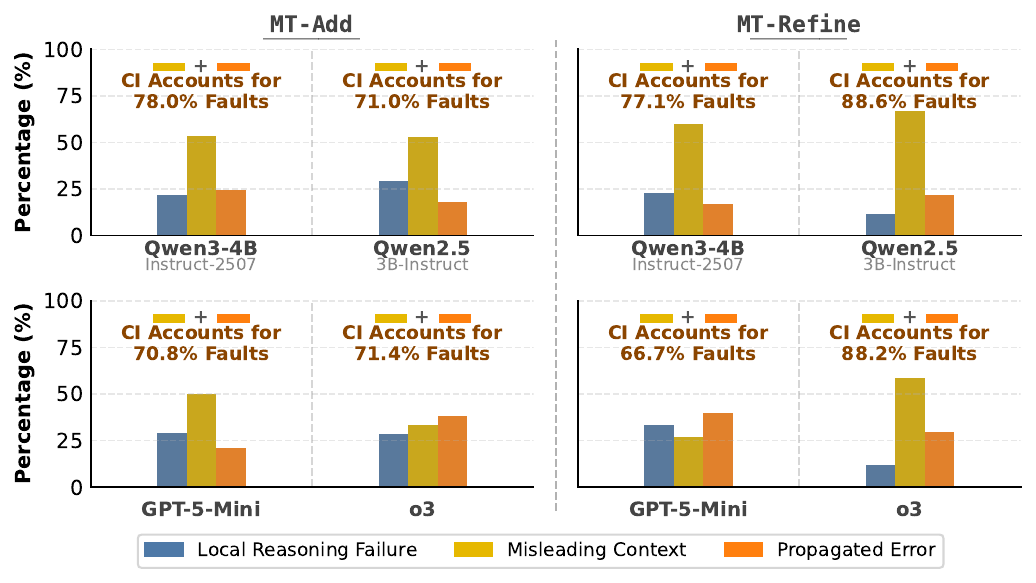}
\caption{Additional experiments: Tracing the root cause of LLM failures.}
\label{fig:root_cause_failures_appdix} 
\end{figure}

We extend our behavioral analysis and root cause tracing to include advanced models, specifically GPT-5-mini \citep{openai2025gpt5} and o3 \citep{openai2025gpto3}. As illustrated in Figure~\ref{fig:Similarity_Vulnerability_appdix} and Figure~\ref{fig:root_cause_failures_appdix}, Indiscriminate nature of contextual inertia persists across different models, even in reasoning-intensive models such as o3. Notably, over 70\% of failures can be traced directly to errors in previous responses. These findings further corroborate our conclusions in Section~\ref{sec:multi_inv}, confirming that Contextual Inertia is the primary driver of LLM vulnerability in multi-turn interactions.

\subsection{Effectiveness without Data Filtering}

\begin{figure}[h]
\centering
\includegraphics[width=0.8\linewidth]{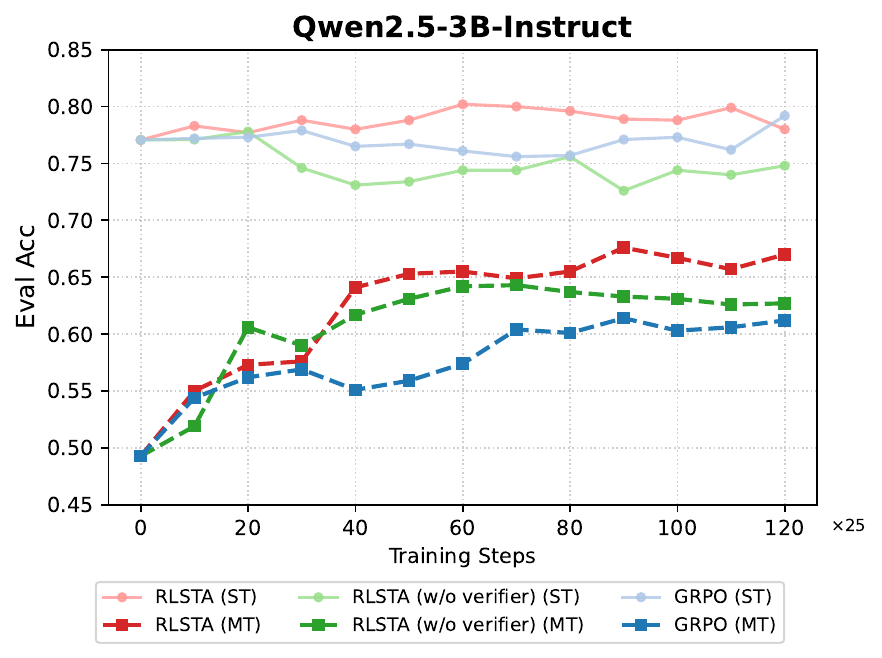}
\caption{Training dynamics of Single-turn (ST) and Multi-turn (MT) performance on datasets without the filtering procedure.}
\label{fig:no_verifier_no_filter} 
\end{figure}

RLSTA leverages the model's own strong reasoning capabilities to provide reward signals for multi-turn responses. While Section~\ref{sec:exp} demonstrated the effectiveness of RLSTA without external verifiers, those experiments utilized filtered data. Here, we conduct additional experiments on datasets without this filtering procedure, as shown in Figure~\ref{fig:no_verifier_no_filter}. Although training on unfiltered data results in lower absolute performance compared to the filtered setting, RLSTA (w/o verifier) still achieves performance comparable to the GRPO baseline on the same data. This further demonstrates the effectiveness and generalizability of RLSTA, particularly in domains where data filtering or external verifiers are not feasible.

\subsection{Generalization to Thinking Models}

\begin{table}[h]
\centering
\caption{Experimental results comparing Base and RLSTA on Qwen3-4B. The model was trained in \textit{No Think} mode but evaluated in both \textit{Think} and \textit{No Think} modes. RLSTA yields consistent improvements across both inference settings.}
\resizebox{\linewidth}{!}{%
\begin{tabular}{lcccccccc}
\toprule

\multirow{3}{*}{\textbf{Method}} & 
\multicolumn{4}{c}{\textbf{Qwen3-4B (No Think)}} & 
\multicolumn{4}{c}{\textbf{Qwen3-4B (Think)}} \\
\cmidrule(lr){2-5} \cmidrule(lr){6-9}
\addlinespace[-2.5pt]

 & \multicolumn{2}{c}{\small \textbf{\texttt{MT-Add}}} & \small \textbf{\texttt{MT-Refine}} & \multirow{2}{*}{ \textbf{Avg}} 
 & \multicolumn{2}{c}{\small \textbf{\texttt{MT-Add}}} & \small \textbf{\texttt{MT-Refine}} & \multirow{2}{*}{ \textbf{Avg}} \\
\cmidrule(lr){2-3} \cmidrule(lr){4-4} 
\cmidrule(lr){6-7} \cmidrule(lr){8-8} 
\addlinespace[-2.5pt]

 & \small Math & \small Code & \small Math & 
 & \small Math & \small Code & \small Math & \\
\midrule

Base & 0.670 & 0.285 & 0.812 & 0.589 & 0.788 & 0.628 & 0.868 & 0.761 \\
\textbf{RLSTA} & \textbf{\underline{0.801}} & \textbf{\underline{0.333}} & \textbf{\underline{0.873}} & \textbf{\underline{0.669}} & \textbf{\underline{0.920}} & \textbf{\underline{0.790}} & \textbf{\underline{0.941}} & \textbf{\underline{0.884}} \\

\bottomrule
\end{tabular}%
}
\label{tab:qwem3_results}
\vspace{-10pt}
\end{table}

Finally, we investigate the applicability of RLSTA to modern reasoning models that utilize ``Thinking'' processes. We select Qwen3-4B \citep{yang2025qwen3} as the base model and train it using RLSTA in \textit{No Think} mode. We then evaluate the model in both \textit{Think} (reasoning enabled) and \textit{No Think} modes during inference.
The results, presented in Table~\ref{tab:qwem3_results}, show that RLSTA significantly enhances performance in both modes. This suggests that RLSTA can effectively transfer the ability of breaking contextual inertia to reasoning-heavy paradigms, even when the training process does not explicitly target the thinking tokens.

\subsection{Single-turn Performance}

\begin{table}[h]
\centering
\caption{Comparison of Base and RLSTA Performance Across Different Models. The $\Delta$ column indicates the relative difference in single-turn performance between Base and RLSTA.}
\vspace{-5pt}
\resizebox{0.6\columnwidth}{!}{
\begin{tabular}{l ccc}
\toprule
\textbf{Model} & \textbf{Base} & \textbf{RLSTA} & \textbf{$\Delta (\%)$} \\
\midrule

\begin{tabular}[l]{@{}l@{}} \small Llama-3.2-3B \\[-0.6ex] {\color{gray}\footnotesize Instruct}\end{tabular} 
& 0.663 & 0.684 & +3.3 \\
\midrule[0.3pt] 

\begin{tabular}[l]{@{}l@{}} \small Qwen2.5-3b \\[-0.6ex] {\color{gray}\footnotesize Instruct}\end{tabular} 
& 0.771 & 0.763 & -0.9 \\
\midrule[0.3pt]

\begin{tabular}[l]{@{}l@{}} \small Qwen2.5-7b \\[-0.6ex] {\color{gray}\footnotesize Instruct}\end{tabular} 
& 0.853 & 0.856 & +0.3 \\
\midrule[0.3pt]

\begin{tabular}[l]{@{}l@{}} \small Qwen3-4B \\[-0.6ex] {\color{gray}\footnotesize Instruct-2507}\end{tabular} 
& 0.894 & 0.898 & +0.4 \\
\midrule[0.3pt]

\begin{tabular}[l]{@{}l@{}} \small Qwen3-4B \\[-0.6ex] {\color{gray}\footnotesize (no think)}\end{tabular} 
& 0.877 & 0.893 & +1.8 \\
\midrule[0.3pt]

\begin{tabular}[l]{@{}l@{}} \small Qwen3-4B \\[-0.6ex] {\color{gray}\footnotesize (think)}\end{tabular} 
& 0.898 & 0.890 & -0.9 \\

\midrule
\textbf{Average} & \textbf{0.826} & \textbf{0.831} & \textbf{+0.7} \\

\bottomrule
\end{tabular}
}
\vspace{-5pt}

\label{tab:st_comparison}
\end{table}

We evaluate the impact of RLSTA on single-turn capabilities on math, as shown in Table \ref{tab:st_comparison}. The results demonstrate that RLSTA preserves the intrinsic single-turn capabilities of the base models while enhancing their multi-turn performance.

\subsection{Contextual Inertia Intensity Using Gemini-2.5-pro}

\begin{figure}[h]
\centering
\vspace{-15pt}
\includegraphics[width=\linewidth]{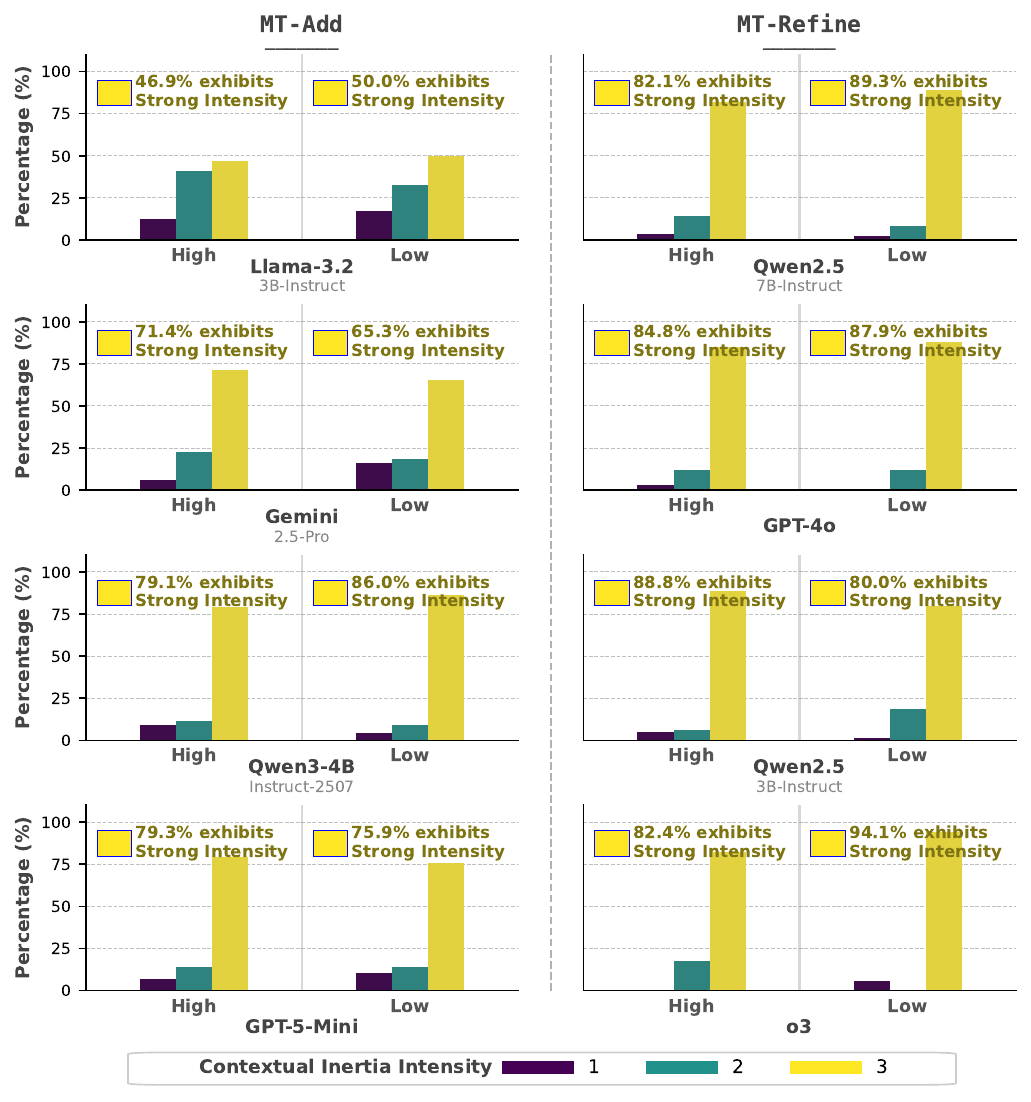}
\vspace{-20pt}
\caption{Distributions of Contextual Inertia Intensity $\texttt{I}_{\mathrm{CI}}(m_n, m_{n-1})$. The inertia intensity are categorized using Gemini-2.5-pro.}
\label{fig:Similarity_Vulnerability_gemini} 
\vspace{-5pt}
\end{figure}

While we mainly analysis the contextual inertia intensity through gpt-4o (Figures~\ref{fig:similarity},\ref{fig:Similarity_Vulnerability_appdix},\ref{fig:Similarity_RLSTA}), here we provide additional analysis using Gemini-2.5-pro \citep{comanici2025gemini}, as shown in Figure~\ref{fig:Similarity_Vulnerability_gemini}, in most cases, the model's final answer $m_n$ still exhibits strong inertia intensity to the preceding response $m_{n-1}$, and the distribution of this intensity remains indistinguishable regardless of the conversation history quality (high vs. low), which align with our observations using gpt-4o, further validate the indiscriminate nature of contextual inertia.

\subsection{Robustness of Latent Capability Filtering across Data Sources}

We demonstrate that our Latent Capability Filtering strategy is not restricted to conversation histories generated by the target model $\mathcal{M}$ itself; it is equally effective when applied to raw histories from external models $\mathcal{M'}$. This transferability stems from the design of our filtering criterion (Eq.~\ref{eq:data_filter}): it relies solely on the \textit{target} model $\mathcal{M}$'s intrinsic single-turn capability. As long as $\mathcal{M}$ possesses the latent knowledge to solve the problem given the full context ($i^{\mathrm{full}}$), it can provide a valid single-turn anchor reward ($R_{\mathrm{s}}$) to guide the multi-turn generation, regardless of the origin of the conversation trajectory.

To validate this, we conduct cross-model experiments where Qwen2.5-3B-Instruct and Llama-3.2-3B-Instruct are trained using raw conversation data generated either by themselves (Self) or by the other model (Cross). The results are presented in Table~\ref{tab:data_source_ablation}.

\begin{table}[h]
\centering
\caption{Cross-model evaluation of Latent Capability Filtering on different data sources. We compare the performance when using raw data generated by Qwen2.5-3B-Instruct versus Llama-3.2-3B-Instruct to train different target models.}
\resizebox{\linewidth}{!}{%
\begin{tabular}{lcccccccc}
\toprule

\multirow{3}{*}{\textbf{Source of $\mathcal D_{\mathrm{raw}}$}} & 
\multicolumn{4}{c}{\textbf{Qwen2.5-3B-Instruct}} & 
\multicolumn{4}{c}{\textbf{Llama-3.2-3B-Instruct}} \\
\cmidrule(lr){2-5} \cmidrule(lr){6-9}
\addlinespace[-2.5pt]

 & \multicolumn{2}{c}{\small \textbf{\texttt{MT-Add}}} & \small \textbf{\texttt{MT-Refine}} & \multirow{2}{*}{ \textbf{Avg}} 
 & \multicolumn{2}{c}{\small \textbf{\texttt{MT-Add}}} & \small \textbf{\texttt{MT-Refine}} & \multirow{2}{*}{ \textbf{Avg}} \\
\cmidrule(lr){2-3} \cmidrule(lr){4-4} 
\cmidrule(lr){6-7} \cmidrule(lr){8-8} 
\addlinespace[-2.5pt]

 & \small Math & \small Code & \small Math & 
 & \small Math & \small Code & \small Math & \\
\midrule

\textbf{Qwen2.5-3B}
 & 0.715 & 0.256 & 0.745 & 0.572 
 & 0.649 & 0.205 & 0.640 & 0.498 \\
\addlinespace[4pt] 

\textbf{Llama-3.2-3B}
 & 0.690 & 0.239 & 0.729 & 0.553 
 & 0.609 & 0.245 & 0.646 & 0.500 \\

\bottomrule
\end{tabular}%
}
\label{tab:data_source_ablation}
\end{table}

As illustrated in Table~\ref{tab:data_source_ablation}, RLSTA exhibits stability across different data sources. Specifically, Qwen2.5-3B-Instruct maintains comparable average performance when trained on Llama-generated histories (0.553) versus its own histories (0.572). Similarly, Llama-3.2-3B shows negligible performance fluctuation (0.498 vs. 0.500). These findings suggest that our approach is robust to the data source, allowing models to effectively break contextual inertia even when learning from conversation trajectories generated by heterogeneous systems.

W2: Simplifying dialogues into just two isolated modes, MT-Add and MT-Refine, may be somewhat biased, as real-world interactions are often an interleaved combination of both.

\subsection{Evaluation on Hybrid Multi-Turn Scenarios}
\label{sec:hybrid_scenarios}

While we primarily focus on \texttt{MT-Add} and \texttt{MT-Refine} as two fundamental multi-turn dialogue paradigms, real-world interactions often involve a complex interplay of both. To further evaluate the robustness of our method in more realistic, interleaved scenarios, we introduce two new hybrid settings based on the math datasets:

\begin{itemize}[nosep, leftmargin=10pt] 
    \item \textbf{Add-then-Refine (AtR):} We simulate a scenario where the user first provides a sequence of corrupted information shards (similar to \texttt{MT-Add} but with noisy data), followed by a sequence of their corresponding corrections. This results in the interaction sequence $[s_1, \dots, s_n, c_1, \dots, c_n]$, representing a user who provides a complete set of initial, flawed details before issuing block corrections.
    
    \item \textbf{Add-Interleaved-with-Refine (AiwR):} We simulate a scenario where the user provides a corrupted shard $s_i$ and immediately follows it with its correction $c_i$ before moving to the next piece of information. This forms the interleaved sequence $[s_1, c_1, s_2, c_2, \dots, s_n, c_n]$, where each element serves as a sequential query. This tests the model's ability to process immediate feedback during continuous information intake.
\end{itemize}

We evaluated our RLSTA-tuned models trained exclusively on standard \texttt{MT-Add} and \texttt{MT-Refine} tasks—against the base models. The results, presented in Table~\ref{tab:hybrid_results}, demonstrate that RLSTA generalizes well to these unseen hybrid structures. Despite not being explicitly trained on interleaved data, RLSTA yields substantial performance improvements over the base models in both AtR and AiwR settings.

\begin{table}[h]
\centering
\caption{Evaluation of Base and RLSTA models on hybrid multi-turn scenarios (Add-then-Refine and Add-Interleaved-with-Refine). Models trained with RLSTA show consistent improvements on unseen hybrid interaction patterns.}
\resizebox{\linewidth}{!}{%
\begin{tabular}{llcc}
\toprule
\textbf{Model} & \textbf{Setting} & \textbf{Base (Acc)} & \textbf{RLSTA (Acc)} \\
\midrule
\multirow{2}{*}{Qwen2.5-3B-Instruct} 
& AtR  & 0.476 & \textbf{0.665} \\
& AiwR & 0.471 & \textbf{0.572} \\
\midrule
\multirow{2}{*}{Qwen2.5-7B-Instruct} 
& AtR  & 0.568 & \textbf{0.794} \\
& AiwR & 0.542 & \textbf{0.600} \\
\midrule
\multirow{2}{*}{Qwen3-4B-2507}       
& AtR  & 0.574 & \textbf{0.858} \\
& AiwR & 0.615 & \textbf{0.863} \\
\bottomrule
\end{tabular}%
}
\label{tab:hybrid_results}
\vspace{-10pt}
\end{table}

\section{Additional Experimental Details}
\label{sec:add_exp_details}

\paragraph{Training Details}
We implement our algorithms using the TRL library \citep{vonwerra2022trl}. For Supervised Fine-Tuning (SFT), we set the learning rate to $1\text{e-}6$ with a batch size of 32, selecting the checkpoint with the lowest validation loss over 5 epochs. For Direct Preference Optimization (DPO), we employ a learning rate of $5\text{e-}7$ with a batch size of 16, training for 1 epoch. For both GRPO and RLSTA, we adopt a learning rate of $3\text{e-}7$, a batch size of 16, KL penalty coefficient 1e-4,  group size of 8, and temperature 1.0.
Regarding training data, we sample 800 questions from the GSM8K dataset \citep{cobbe2021training} and rewrite them into multiple shards. For data generation, following \citet{Peng2025LargeRM}, in Section~\ref{sec:exp}, we utilize Qwen2.5-3B-Instruct to first generate raw multi-turn conversation histories during training, which are subsequently filtered specifically for each model. We consider both the \texttt{MT-Add} and \texttt{MT-Refine} scenarios. Specifically, for the Qwen series models (including Qwen2.5-3B/7B-Instruct and Qwen3-4B-Instruct-2507 / Qwen3-4B), we filter and retain 100 \texttt{MT-Refine} and 400 \texttt{MT-Add} conversation histories. For Llama-3.2-3B-Instruct, we retain 200 \texttt{MT-Refine} and 400 \texttt{MT-Add} conversation histories.

\paragraph{Evaluation Details}
Our evaluation follows the protocol of \citet{laban2025llms}. We run 8 simulations per instruction with a temperature of 0.7, reporting the average performance. The maximum token limit is set to 1,024 for No Think models and 8,192 for Qwen3-4B in thinking mode. We extract the final answer from the response using GPT-4o-mini for Math and Action tasks, while employing rule-based extraction for Code and Database tasks. During training, to reduce costs, we employ a simplified extraction strategy: we identify the last four numbers in the response and check whether the correct answer is present among them. Our results demonstrate that this simplified method effectively teaches the model to provide the correct final answer.
\section{Case Studies}
In this section, we present a comparative case study between the base model and the RLSTA-aligned model. Both models are presented with an identical multi-turn conversation history containing a misleading intermediate reasoning trace. As illustrated below, the base model exhibits Contextual Inertia: it is misled by its own previous response, rigidly adhering to the hallucinated condition that 'insurance covers 80\% of the cost,' which ultimately propagates into an incorrect final answer. In contrast, the RLSTA-tuned model successfully breaks this inertia. It filters out the invalid context from the previous turn, recalculates based solely on the user's verified constraints, and derives the correct total cost.

\begin{tcolorbox}[
    breakable, 
    title={\textbf{Case Study for Base model}}, 
    colback=white,       
    colframe=black!75,   
    boxrule=0.8pt,       
    arc=0mm,             
]
{\small
\textbf{\textit{Previous Information (Users):}}
\begin{itemize}[leftmargin=*, nosep, topsep=2pt] 
    \item Calculate the total cost of John's physical therapy?
    \item John hurt his back and needed physical therapy.
    \item He attended therapy for 6 weeks, twice per week.
    \item Each session lasted 2 hours.
\end{itemize}

\vspace{0.5em}
\hrule 
\vspace{0.5em}

\textbf{\textit{Last Assistant Response:}}

\noindent \textbf{Assistant:} Given that each session lasted 2 hours, we need to adjust our calculations to account for the duration of each session. 

[...] 

\textcolor{magenta}{if the insurance covers 80\% of the total cost, then John only pays 20\% of the total cost.}

[...]

\vspace{0.5em}
\hrule 
\vspace{0.5em}

\textbf{\textit{Final Interaction:}}

\vspace{0.3em}
\noindent \textbf{User:} the cost per hour for a session was \$125.

\noindent \textbf{Assistant:} Given that the cost per hour for a session is \$125, we need to adjust our calculations to account for the duration of each session.

[...] 

\noindent \textcolor{red}{Insurance covers 80\%, so John pays 20\%. Total cost after insurance : 0.2 times 3000 = 600}

\noindent Therefore, the total cost of John's physical therapy after insurance coverage is \textcolor{red}{\boxed{600}}
}
\end{tcolorbox}

\begin{tcolorbox}[
    breakable, 
    title={\textbf{Case Study for Model after RLSTA}}, 
    colback=white,       
    colframe=black!75,   
    boxrule=0.8pt,       
    arc=0mm,             
]
{\small
\textbf{\textit{Previous Information (Users):}}
\begin{itemize}[leftmargin=*, nosep, topsep=2pt] 
    \item Calculate the total cost of John's physical therapy?
    \item John hurt his back and needed physical therapy.
    \item He attended therapy for 6 weeks, twice per week.
    \item Each session lasted 2 hours.
\end{itemize}

\vspace{0.5em}
\hrule 
\vspace{0.5em}

\textbf{\textit{Last Assistant Response:}}

\noindent \textbf{Assistant:} Given that each session lasted 2 hours, we need to adjust our calculations to account for the duration of each session. 

[...] 

\textcolor{magenta}{if the insurance covers 80\% of the total cost, then John only pays 20\% of the total cost.}

[...]

\vspace{0.5em}
\hrule 
\vspace{0.5em}

\textbf{\textit{Final Interaction:}}

\vspace{0.3em}
\noindent \textbf{User:} the cost per hour for a session was \$125.

\noindent \textbf{Assistant:} Let's solve the problem step by step.

[...] 

\noindent Perform the multiplication: 24 times 125 = 24 times (100 + 25) = (24 times 100) + (24 times 25) 24 times 100 = 2400 24 times 25 = 600 2400 + 600 = 3000

\noindent Therefore, the total cost of John's physical therapy is \textcolor{green}{\boxed{3000}}
}
\end{tcolorbox}
\section{Prompts}

\subsection{Similarity Analysis}
\definecolor{simLow}{HTML}{440154} 
\definecolor{simMid}{HTML}{21918c} 
\definecolor{simHigh}{HTML}{D4C000}

We employ GPT-4o to assess the logical similarity between response pairs, focusing specifically on their reasoning structures and key derivation steps. We quantify this similarity on a 5-point scale. For the visualizations presented in Figure~\ref{fig:similarity}, Figure~\ref{fig:Similarity_RLSTA}, and Figure~\ref{fig:Similarity_Vulnerability_appdix}, we categorize these scores into three distinct tiers represented by a color gradient: \textcolor{simHigh}{\textbf{High Similarity}} (Levels 4--5), \textcolor{simMid}{\textbf{Medium Similarity}} (Level 3), and \textcolor{simLow}{\textbf{Low Similarity}} (Levels 1--2).

\lstset{
    language=Python,                
    basicstyle=\ttfamily\scriptsize, 
    breaklines=true,                
    breakatwhitespace=false,        
    frame=single,                   
    rulecolor=\color{black!30},     
    keywordstyle=\color{blue},      
    stringstyle=\color{green!50!black}, 
    commentstyle=\color{gray},      
    showstringspaces=false,         
    captionpos=b,                   
    escapeinside={(*@}{@*)}         
}

\begin{lstlisting}
EVALUATION_SYSTEM_PROMPT = """You are a Senior Logic Analyst specializing in auditing multi-turn AI reasoning and Chain-of-Thought (CoT) consistency.

Your objective is to evaluate how well an AI assistant adapts its logic to changing constraints while maintaining mathematical precision. You must perform a deep structural comparison between the provided Example, First Turn, and Second Turn.

Key Responsibilities:
**Verify Accuracy:** Rigorously check the final answer against the Ground Truth.

Output your analysis in strict JSON format. Do not include markdown formatting (like ```json) or conversational text."""

EVALUATION_USER_PROMPT_TEMPLATE = """
## Inputs

### Example Answer Attempt (example_answer_attempt):
{example_answer_attempt}

### First Turn
**Query (first_turn_query):**
{first_turn_query}

**Assistant Response (first_turn_response):**
{first_turn_response}

### Second Turn
**Query (second_turn_query):**
{second_turn_query}

**Assistant Response (second_turn_response):**
{second_turn_response}

### Ground Truth
**Correct Answer (correct_answer):**
{correct_answer}

---

## Evaluation Instructions

### Multi-Turn Response Similarity Assessment
Analyze the similarity between responses, focusing on logic structure and Chain of Thought (CoT) steps. Provide a similarity score (1-5) for the following pairs:
1. **example-r1**: Similarity between `example_answer_attempt` and `first_turn_response`.
2. **example-r2**: Similarity between `example_answer_attempt` and `second_turn_response`.
3. **r1-r2**: Similarity between `first_turn_response` and `second_turn_response`.

**Similarity Score Definitions:**
- **1 (Distinct):** The response uses a completely fresh approach; no apparent similarity in logic or structure.
- **2 (Minimal):** The response is mostly independent; shares only superficial elements but uses different reasoning.
- **3 (Moderate):** The response reuses some steps or general methods but includes meaningful adaptations to the CoT structure.
- **4 (Strong):** The response heavily reuses the CoT structure and steps, with limited adaptation (e.g., simply changing numbers while keeping the exact same logic flow).
- **5 (Near-identical):** The response mirrors the reference with minimal changes, using almost the same logic structure, steps, and phrasing.
---

## Output Format

Return exactly one JSON object matching this schema:

```json
{{
  "similarity_assessment": {{
  "analysis process - example-r1": "<Detailed explanation of the similarity assessment>",
  "example-r1": 1 | 2 | 3 | 4 | 5,
  "analysis process - example-r2": "<Detailed explanation of the similarity assessment>",
  "example-r2": 1 | 2 | 3 | 4 | 5,
  "analysis process - r1-r2": "<Detailed explanation of the similarity assessment>",
  "r1-r2": 1 | 2 | 3 | 4 | 5
  }},
  "notes": "string"
}}

Return only the JSON object with no additional text or markdown formatting.
"""
\end{lstlisting}

\subsection{Data Preparation}

We directly adapt the prompts in \citet{laban2025llms} to process a single-turn instruction in GSM8K into shards. The specific prompts used for segmentation and rephrasing are provided below:

\lstset{
    language=Python,                
    basicstyle=\ttfamily\scriptsize, 
    breaklines=true,                
    breakatwhitespace=false,        
    frame=single,                   
    rulecolor=\color{black!30},     
    keywordstyle=\color{blue},      
    stringstyle=\color{green!50!black}, 
    commentstyle=\color{gray},      
    showstringspaces=false,         
    captionpos=b,                   
    escapeinside={(*@}{@*)}         
}

\begin{lstlisting}
Segmentation_Prompt_Format = """
You are a given a arithmetic question, and your task is to segment the question into units of information that each reveal a single piece of information of the question.
You must output a list of segments in the following JSON format:
[
    {"segment": "[exact excerpt from the question]", "is_required": 1|0},
    {"segment": "[exact excerpt from the question]", "is_required": 1|0},
    ...
]

Rules:
- [is_required] For each segment, you must specify whether this particular segment is required (1) or not strictly necessary (0) to answer the question.
- [Non-overlapping] The segments must be non-overlapping and cover the entire question. You can optionally leave some gaps for non-essential portions of the original question (delimiters, headers, etc.)
- [Minimalistic] You should split the information in the segments to as small as possible. If you have a compound expression (X and Y), you should split it into two segments. Each segment should represent a unit of information.
- [Valid Segments] Only extract segments from the text of the question. Do not include example inputs/outputs as segments.
- [Segment count] The number of segments should not be more than 10.

Example Question:

Q: There are 15 trees in the grove. Grove workers will plant trees in the grove today. After they are done, there will be 21 trees. How many trees did the grove workers plant today?


Output:
{"segments": [
    {"segment": "There are 15 trees in the grove", "is_required": 1},
    {"segment": "Grove workers will plant trees in the grove today", "is_required": 1},
    {"segment": "After they are done, there will be 21 trees", "is_required": 1},
    {"segment": "How many trees did the grove workers plant today?", "is_required": 1},
]}

Now complete the task for the following fully specified question:

"""

Rephrasing_Prompt_Format = """
You are given a segment of a complete question, and your task is to: (1) choose one that will be the initial query of a multi-step query, and then each of the remaining segment should be one int provided to the system in a follow-up turn of the conversation.

Your output should be a JSON object in the following format:
{
    "initial_segment": "[exact excerpt from the question]",
    "initial_query": "conversational version of the initial segment",
    "hints": [
    {"segment": "[exact excerpt from the question]", "hint": "conversational version of the segment taking the rest of the question into account"}
    ]
}

Example:

Q: There are 15 trees in the grove. Grove workers will plant trees in the grove today. After they are done, there will be 21 trees. How many trees did the grove workers plant today?

Segments:
[
    {"segment": "There are 15 trees in the grove", "is_required": 1},
    {"segment": "Grove workers will plant trees in the grove today", "is_required": 0},
    {"segment": "After they are done, there will be 21 trees", "is_required": 1},
    {"segment": "How many trees did the grove workers plant today?", "is_required": 1},
]

Output:
{
    "initial_segment": "How many trees did the grove workers plant today?",
    "initial_query": "I need to calculate the number of trees planted by the grove workers today",
    "hints": [
        {"segment": "There are 15 trees in the grove", "hint": "15 trees are in the grove before planting"},
        {"segment": "Grove workers will plant trees in the grove today", "hint": "Grove workers will plant trees in the grove today"},
        {"segment": "After they are done, there will be 21 trees", "hint": "I see 21 trees after the grove workers are done"},
    ]
}

Rules:
- [Query selection] Choose the question segment from the segments to form the initial query.
- [Transform each segment] Make sure each segment is included either as the initial query or as a hint. Do not forget any segments.
- [Short initial query] Make the initial query short, not a full sentence, similar to how users use a search engine like Google
- [Order of hints] Order the hints in order of importance, from most to least important to the query. You do not need to keep the order the segments are provided in.

Now complete the task for the following fully specified question and segments:

"""
\end{lstlisting}

\subsection{Abstain Prompt}
While our standard evaluation follows the default system prompt protocol outlined in \citet{laban2025llms}, we introduce a variant setting to facilitate a direct comparison with RLAAR, an abstention-based strategy. Specifically, we evaluate our method with an additional system instruction that explicitly encourages the model to withhold answers when information is incomplete (denoted as RLSTA + Abstain). The specific prompt used for this setting is provided below:
\begin{lstlisting}
[Important] I will provide additional conditions incrementally in turns rather than all at once. Do not provide a concrete answer attempt you have all necessary details.
After each new condition, review your previous response to ensure it remains correct and compatible.
Update your solution accordingly while maintaining correctness under all conditions provided so far.
\end{lstlisting}

\subsection{Corrupting Prompt}

In the \texttt{MT-Refine} task, we employ GPT-4o to systematically alter specific details in the information shards to corrupt the initial query. The prompt used for this process is as follows:
\begin{lstlisting}
prompt = f"""
You are a data augmentation expert.
Below is a list of "shards" representing parts of a math problem.

Your Task:
1. I have excluded the first shard (the main question).
2. You must modify **EVERY** single shard provided in the list below.
3. Apply the following modification style:
    **STRICT NUMERIC MODIFICATION**:
    - Identify the numbers/integers in the text.
    - Change the values of these numbers to different reasonable values.
    - Do NOT change the words, units (years, fruits), or the logic (multiplication, addition).
    - ONLY change the digits.

Input Data (Shards to modify):
{shards_text}

Output Format:
Return a JSON object with a single key "modified_shards" containing the list of modified objects.
Example:
{{
  "modified_shards": [
      {{ "shard_id": 2, "shard": "modified text..." }},
      {{ "shard_id": 3, "shard": "modified text..." }}
  ]
}}
"""
\end{lstlisting}

\end{document}